\documentclass[runningheads]{llncs}
\makeatletter
\renewcommand\subsubsection{\@startsection{subsubsection}{3}{\z@}%
                       {-18\p@ \@plus -4\p@ \@minus -4\p@}%
                       {0.5em \@plus 0.22em \@minus 0.1em}%
                       {\normalfont\normalsize\bfseries\boldmath}}
\makeatother
\usepackage{graphicx}
\graphicspath{{./}}
\usepackage[T1]{fontenc}
\usepackage[utf8]{inputenc}
\usepackage{graphicx}
\usepackage{rotating}
\usepackage{lscape}
\usepackage{float}
\usepackage{multirow}
\usepackage{array}
\usepackage{footnote}
\usepackage{url}
\usepackage{supertabular}
\usepackage{longtable}
\usepackage{amsmath}
\usepackage{xcolor}
\usepackage{ulem}

\newcolumntype{P}[1]{>{\centering\arraybackslash}p{#1}}
\newcommand{\comment}[1]{}

\begin{document}


\title{Biologically Plausible Learning of Text Representation with Spiking Neural Networks}
\author{Marcin Białas\inst{1}\orcidID{0000-0001-8121-7401} \and
Marcin Michał Mirończuk\inst{1}\orcidID{0000-0002-4951-2264} \and
Jacek Ma{\'n}dziuk\inst{2}\orcidID{0000-0003-0947-028X}}
\authorrunning{Marcin Białas et al.}
\titlerunning{Spike Encoder for Text}
\institute{
    National Information Processing Institute, al. Niepodległości 188 b, 00-608 Warsaw, Poland,
    \email{\{marcin.bialas,marcin.mironczuk\}@opi.org.pl}
    \and
   Faculty of Mathematics and Information Sciences, Warsaw University of Technology, Koszykowa 75, 00-662 Warsaw, Poland, \tt
    \email{mandziuk@mini.pw.edu.pl}
    }

\maketitle

\begin{abstract}
This study proposes a novel biologically plausible mechanism for generating low-dimensional spike-based text representation. First, we demonstrate how to transform documents into series of spikes (\textit{spike trains}) which are subsequently used as input in the training process of a spiking neural network (SNN). The network is composed of biologically plausible elements, and trained according to the unsupervised Hebbian learning rule, Spike-Timing-Dependent Plasticity (STDP). After training, the SNN can be used to generate low-dimensional spike-based text representation suitable for text/document classification. Empirical results demonstrate that the generated text representation may be effectively used in text classification leading to an accuracy of $80.19\%$ on the \textit{bydate} version of the \textit{20 newsgroups} data set, which is a leading result amongst approaches that rely on low-dimensional text representations.

\keywords{spiking neural network \and STDP \and Hebbian learning \and text processing \and text representation \and spike-based representation \and representation learning \and feature learning \and text classification \and 20 newsgroups bydate
}
\end{abstract}

\section{Introduction}
Spiking neural networks (SNNs) are an example of biologically plausible artificial neural networks (ANNs). SNNs, like their biological counterparts, process sequences of discrete events occurring in time, known as spikes. Traditionally, spiking neurons, due to their biological validity, have been studied mostly by theoretical neuroscientists, and have become a standard tool for modeling brain processes on a micro scale. However, recent years have shown that spiking computation can also successfully address common machine learning challenges~\cite{Tavanaei2019}. Another interesting aspect of SNNs is the adaptation of such algorithms to neuromorphic hardware which is a brain-inspired alternative to the traditional von Neumann machine. Thanks to mimicking processes observed in brain synaptic connections, neuromorphic hardware is a highly fault-tolerant and energy-efficient substitute for classical computation~\cite{nawrocki:voleys}.

Recently we have witnessed significant growth in the volume of research into SNNs. Researchers have successfully adapted SNNs for the processing of images~\cite{Tavanaei2019}, audio signals~\cite{DominguezMorales2018,Wu2018,Wysoski2010}, and time series~\cite{Kasabov2015,Reid2014}. However, to the best of the authors knowledge, there is only one work related to text processing with SNNs~\cite{Wang2019}. This state of affairs is caused by the fact that text, due to its structure and high dimensionality, presents a significant challenge to tackle by the SNN approach. The motivation of this study is to broaden the current knowledge of the application of SNNs to text processing. More specifically, we have developed and evaluated a novel biologically inspired method for generation of spike-based text representation that may be used in text/document classification task~\cite{Mladenic2017}.

\subsection{Objectives and summary of approach} This paper proposes an \textit{Spike Encoder for Text} (SET) which generates spike-based text representation suitable for classification task. Text data is highly dimensional (the most common text representation is in the form of a vector with many features) which, due to the \textit{curse of dimensionality}~\cite{Keogh2017,Aggarwal15,James2013,Murphy2012}, usually leads to overfitted classification models with poor generalisation~\cite{Webb2017,SILVA2017152,RAZA2019341,HARTMANN201920,ASIF2020101345}.

Processing highly dimensional data is also computationally expensive.
Therefore, researchers have sought text representations which may overcome this drawback~\cite{Aggarwal2018}. One of possible approaches is based on transformation of high dimensional feature space to low-dimensional representation~\cite{AYESHA202044,Vlachos2017,Bengio2013}.

In the above context we propose the following two-phase approach to SNN based text classification. Firstly, the text is transformed into \textit{spike trains}. Secondly, spike trains representation is used as the input in the SNN training process performed according to biologically plausible unsupervised learning rule, and generating the spike-based text representation. This representation has significantly lower dimensionality than the spike trains representation and can be used effectively in subsequent SNN text classification. The proposed solution has been empirically evaluated on the publicly available version, \textit{bydate}~\cite{Lang95} of the real data set known as \textit{20 newsgroups}, which contains $18\,846$ text documents from twenty different newsgroups of \textit{Usenet}, a worldwide distributed discussion system.

Both the input and output of the SNN rely on spike representations, though of very different forms. For the sake of clarity, throughout the paper the former representation (SNN input) will be referred to as \textit{spike trains}, and the latter one (SNN output) as  \textit{spike-based}, or \textit{spiking encoding}, or \textit{low-dimensional}.

\subsection{Contribution} The main contribution of this work can be summarized as follows:
\begin{itemize}
\item To propose an original approach to document processing using SNNs and its subsequent classification based on generated  spike-based text representation;
\item To experimentally evaluate the influence of various parameters on the quality of generated representation, which leads to better understanding of the strengths and limitations of SNN-based text classification approaches;
\item To propose an SNN architecture which may potentially contribute to development of other SNN based approaches. We believe that the solution presented may serve as a building block for larger SNN architectures, in particular deep spiking neural networks (DSNNs)~\cite{Tavanaei2019};
\end{itemize}

\subsection{Related work}\label{sec:rw}
As mentioned above, we are aware of only one paper related to text processing in the context of SNNs context~\cite{Wang2019} which, nevertheless, differs significantly from our approach. The authors of~\cite{Wang2019} focus on transforming word embeddings~\cite{Mikolov2013,Pennington2014} into spike trains,
whilst our focus is not only on representation of text in the form of spike trains, but also on training the SNN encoder which generates l low-dimensional text representation. In other words, our goal is to generate a low-dimensional text representation with the use of SNN base, whereas in~\cite{Wang2019} the transformation of an existing text embedding into spike trains is proposed.

This remainder of the paper is structured as follows.
Section~\ref{sec:method} presents an overview of the proposed method; Section~\ref{sec:experiment} describes the evaluation process of the method and experimental results; and Section~\ref{sec:conclusions} presents the conclusions.

\section{Proposed spiking neural method}\label{sec:method}

\begin{figure}[h]
\includegraphics[width=\textwidth]{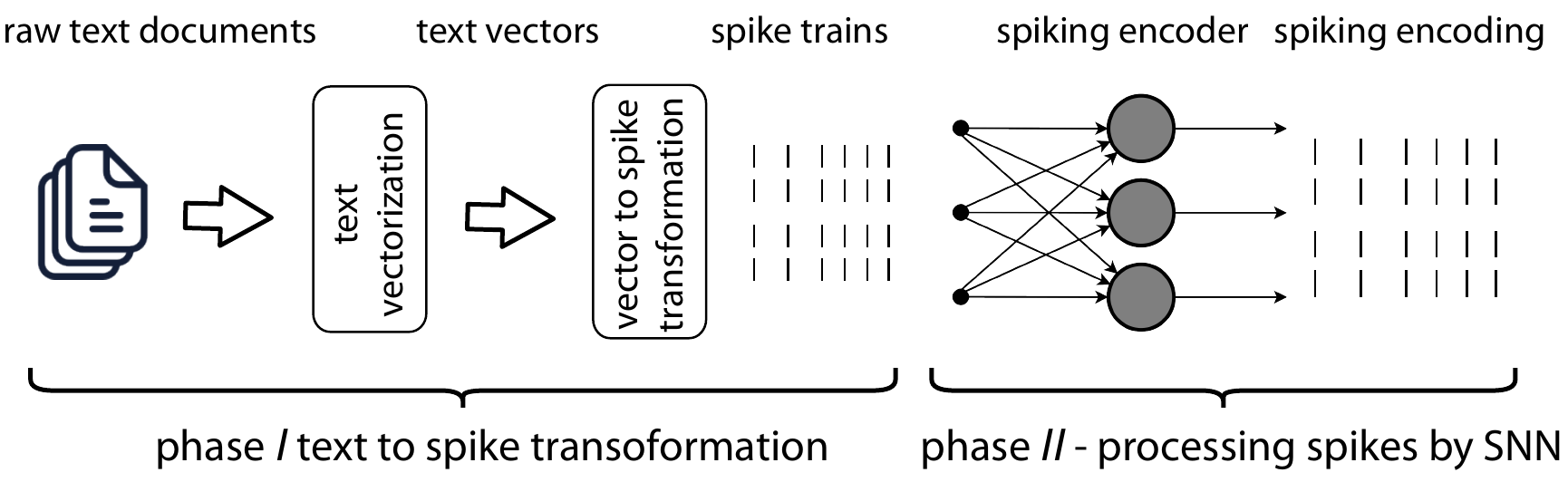}
\caption{A schema of the proposed method for generating spike-based low-dimensional text representation.}
\label{fig:proposedMethod}
\end{figure}

The proposed method transforms input text to spike code and uses it as training input for the SNN to achieve a meaningful spike-based text representation. The method is schematically presented in Fig.~\ref{fig:proposedMethod}.
In phase \textit{I}, text is transformed into a vector representation and afterwards each vector is encoded as spike trains.
Once the text is encoded in the form of neural activity, it can be used as input to the core element of our method - \textit{a spiking encoder}. The encoder is a two-layered SNN with adaptable synapses. During the learning phase (\textit{II}), the spike trains are propagated through the encoder in a feed-forward manner  and synaptic weights are modified simultaneously according to unsupervised learning rule. After the learning process, the output layer of the spiking encoder provides spike-based representation of the text presented to the system.

In the remainder of this section all elements of the system described above are discussed in more detail.

\subsection{Input transformation}\label{sec:inputEncoding}

\subsubsection{Text vectorization}
During a \textit{text to spike transformation} phase like the one illustrated in Fig.~\ref{fig:proposedMethod} text is preprocessed for further spiking computation. Text input data (\textit{data corpus}) is organized as a set $D$ of documents $d_i, i=1,\ldots,K$. In the first step a dictionary $T$ containing all unique words $t_j, j=1,\ldots,|T|$ from the corpus data is built. Next, each document $d_{i}$ is transformed into an $M$-dimensional ($M = |T|$) vector $W_i$, the elements of which, $W_i[j]:=w_{ij}, j=1,\ldots,M$ represent the relevance of words $t_j$ to document $d_i$. In effect, the corpus data is represented by a real-valued matrix $W_{K\times M}$ also called ~\textit{document-term matrix}. 

The typical weighting functions are \textit{term-frequency} (TF), \textit{inverse document frequency} (IDF), or their combination TF-IDF~\cite{Manning2008,Haddoud2016}.
In TF the weight $w_{ij}$ is equal to the number of times the $j$-th word appears in $d_i$ with respect to the length of $|d_i|$ (the number of all non-unique words in $d_i$). IDF takes into account the whole corpus $D$ and sets $w_{ij}$ as the logarithm of a ratio between $|D|$ and the number of documents containing word $t_j$. Consequently, IDF mitigates the impact of words that occur very frequently in a given corpus and are presumably less informative from the point of view of document classification than the words occurring in a small fraction of the documents. TF-IDF sets $w_{ij}$ as a product of TF and IDF weights. In this paper we use TF-IDF weighting which is the most popular approach in text processing domain.

\subsubsection{Vector to spike transformation}
In order to transform a vector representation to spike trains one, presentation time $t_p$ which establishes for how long each document is presented to the network, and the time gap between two consecutive presentations $\Delta t_p$, must be defined. A time gap period, without any input stimuli is necessary to eliminate interference between documents and allow dynamic parameters of the system to decay and  ``be ready'' for the next input.

Technically, for a given document $d_i$, represented as $M$ dimensional vector of weights $w_{ij}$, for each weight $w_{ij}$ in every millisecond of document presentation a spike is generated with probability proportional to $w_{ij}$. Thanks to this procedure, we ultimately derive a spiking representation of the text.

In our experiments each document is presented for $t_p=600[ms]$ and $\Delta t_p=300[ms]$, and proportionality coefficient $\alpha$ is set to $1.5$.

For a better clarification, let's consider a simple example and assume that for a word \textit{baseball} the corresponding weight $w_{ij}$ in some document $d_i$ is equal to $0.1$. Then for each millisecond of a presentation time a probability of emitting a spike $P(spike|baseball)$ equals $\alpha \cdot 0.1 = 0.15$. Hence, $90$ spikes during $600[ms]$ presentation time are expected, on average, to be generated.

\subsection{Spiking encoder architecture and dynamics}\label{sec:inputEncoding}
\begin{figure}[h]
\includegraphics[width=\textwidth]{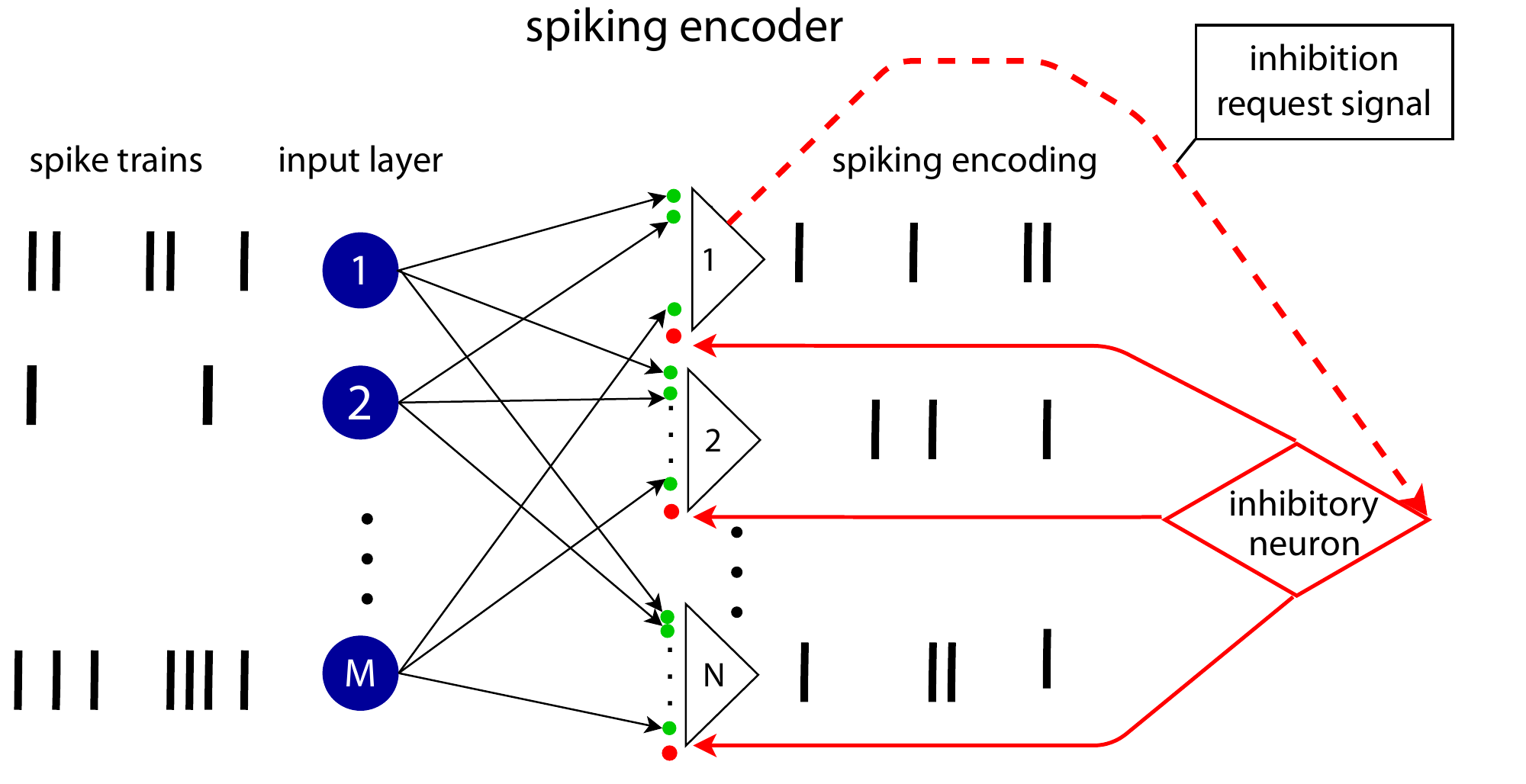}
\caption{Spiking encoder architecture.}
\label{architecture_fig}
\end{figure}

A spiking encoder is the key element of the proposed method. The encoder, presented in Fig.~\ref{architecture_fig}, is a two layered SNN equipped with an additional inhibitory neuron.
The first layer contains $M$ neurons (denoted by blue circles) and each of them represents one word $t_{j}$ from the dictionary $T$. Neuron dynamics is defined by the \textit{spike trains} generated based on weights $w_{ij}$ corresponding to documents $d_i, i=1,\ldots,K$.
Higher numbers of spikes are emitted by neurons representing words which are statistically more relevant for a particular document,  according to the chosen TF-IDF measure. The \textit{spike trains} for each neuron are presented in Fig.~\ref{architecture_fig} as a row of short vertical lines.

In the brain spikes are transmitted between neurons via synaptic connections. A neuron which generates a spike is called a \textit{presynaptic neuron}, whilst a target neuron (spike receiver) is a \textit{postsynaptic neuron}. In the proposed SNN architecture (cf. Fig.~\ref{architecture_fig}) two different types of synaptic connections are utilised: \textit{excitatory} ones and  \textit{inhibitory} ones. Spikes transmitted through excitatory connections (denoted by green circles in Fig.~\ref{architecture_fig}) leads to firing of postsynaptic neuron, while impulses traveling through inhibitory ones (red circles in Fig.~\ref{architecture_fig}) hinder \textit{postsynaptic neuron} activity. Each time an encoder neuron fires its weights are modified according to the proposed learning rule. The neuron simultaneously sends an \textit{inhibition request signal} to the \textit{inhibitory neuron} and activates it. Then the inhibitory neuron suppresses the activity of all encoder output layer neurons using \textit{recursive inhibitory connection} (red circles). The proposed architecture satisfies the competitive learning paradigm~\cite{kaski:kohonen} with a winner-takes-all (WTA) strategy.

In this work we consider a biologically plausible neuron model known as leaky integrate and fire (LIF)~\cite{Gerstner2002}. The dynamics of such a neuron is described in terms of changes of its membrane potential (MP). If the neuron is not receiving any spikes its potential is close to the value of $u_{rest}=-65[mV]$ known as resting membrane potential. When the neuron receives spikes transmitted through excitatory synapses, the MP moves towards excitatory equilibrium potential, $u_{exc}=0[mV]$. When many signals are simultaneously transmitted through excitatory synapses the MP rises and at some point can reach a threshold value of $u_{th}=-52[mV]$, in which case the neuron fires. After firing, the neuron resets its MP to $u_{rest}$ and becomes inactive for $t_{ref}=3[ms]$ (the refractory period). In the opposite scenario, when the neuron receives spikes through the inhibitory synapse, its MP moves towards inhibitory equilibrium potential $u_{inh}=-90[mV]$, i.e. further away from its threshold value, which decreases the chance of firing. The dynamics of the membrane potential $u$ in the LIF model is described by the following equation:
\begin{equation}
\tau\frac{du}{dt}=(u_{rest}-u)+g_{e}(u_{exc}-u)+g_{i}(u_{inh}-u)
\end{equation}
where $g_{e}$ and $g_{i}$ denote excitatory and inhibitory conductance, resp. and $\tau=100[ms]$ is membrane time constant. The values of $g_{e}$ and $g_{i}$ depend on presynaptic activity. Each time a signal is transmitted through the synapse the conductance is incremented by the value of weight corresponding to that synapse, and decays with time afterwards according to equation (\ref{eq:decay})
\begin{equation}
\tau_{e}\frac{dg_{e}}{dt}=-g_{e}  ,\quad \tau_{i}\frac{dg_{i}}{dt}=-g_{i}
\label{eq:decay}
\end{equation}
, where $\tau_{e}=2[ms]$, $\tau_{i}=2[ms]$ are decay time constants. In summary, if there is no presynaptic activity the MP converges to $u_{rest}$. Otherwise, its value changes according to the signals transmitted through the neuron synapses.
\subsection{Hebbian synaptic plasticity}\label{learningRule}
We utilise a modified version of the Spike-Timing-Dependent Plasticity (STDP) learning process~\cite{song:abbot}. STDP is a biologically plausible unsupervised learning protocol belonging to the family of Hebbian learning (HL) methods~\cite{hebb:rule}. In short, the STDP process results in an increase of the synaptic weight if the postsynaptic spike is observed soon after presynaptic one ('pre-before-post'), and in a decrease of the synaptic weight in the opposite scenario ('post-before-pre'). The above learning scheme increases the relevance of those synaptic connections which contribute to the activation of the postsynaptic neuron, and decreases the importance of the ones which do not. We modify STDP in a manner similar to~\cite{diehl:cook,querlioz:bichler}, i.e. by skipping the weight modification in the post-before-pre scenario and introducing an additional scaling mechanism. The plasticity of the excitatory synapse $s_{ij}$ connecting a presynaptic neuron $i$ from the input layer with postsynaptic neuron $j$ from the encoder layer can be expressed as follows:
\begin{equation}
\Delta s_{ij} = \eta(A(t)-(R(t)+0.1)s_{ij})
\label{eq:deltasij}
\end{equation}
where
\begin{equation}
A(t)=-\tau_{A}\frac{dA(t)}{dt},\quad R(t)=-\tau_{R}\frac{dR(t)}{dt}
\label{eq:decay}
\end{equation}
and $\eta=0.01$ is a small learning constant. In eqs. (\ref{eq:deltasij})-(\ref{eq:decay}) $A(t)$ represents a presynaptic trace and $R(t)$ is a scaling factor which depends on the history of postsynaptic neuron activity. Every time the presynaptic neuron $i$ fires $A(t)$ is set to $1$ and exponentially decays in time ($\tau_{A}=5[ms]$). If the postsynaptic neuron fires just after the presynaptic one ('pre-before-post') $A(t)$ is close to $1$ and the weight increase is high. The other component of eq. (\ref{eq:deltasij}) $(R(t)+0.1)s_{ij}$ is a form of synaptic scaling~\cite{miller}. Every time the postsynaptic neuron fires $R(t)$ is incremented by 1 and afterwards decays with time ($\tau_{R}=70[ms]$). The role of the small constant factor $0.1$, is to maintain scaling even when activity is relatively small. The overall purpose of synaptic scaling is to decrease the weights of the synapses which are not involved in firing the postsynaptic neuron. Another benefit of synaptic scaling is to restrain weights from uncontrolled growth which can be observed in HL~\cite{abbot:nelson}.

\subsection{Learning procedure}\label{learning_procedure}
{For a given \textit{data corpus} (set of documents) $D$} the training procedure is performed as follows. Firstly, we divide $D$ into $s$ subsets $u_{i}, i=1,\ldots,s$ in the manner described in Section~\ref{sec:experimentplan}. Secondly, each subset $u_{i}$ is transformed to spike trains and used as input for a separate SNN encoder $H_{i}, i=1,\ldots,s$ composed of $N$ neurons. Please note that
each encoder is trained with the use of one subset only.
Such a training setup allows the processing of the data in parallel manner. Another advantage is that this limits the number of excitatory connections per neuron, which reduces computational complexity (the number of differential equations that need to be evaluated for each spike) as during training encoder $H_{i}$ is exposed only to the respective subset, $T_{i}$ of the training set dictionary $T$ and the number of its excitatory connections is limited to $|T_{i}|<|T|$. Spike trains are presented to the network four times (in four training epochs).

Once the learning process is completed, the \textit{connection pruning} procedure is applied. Please observe that HL combined with competitive learning should lead to highly specialised neurons which are activated only for some subset of the inputs. The specialisation of a given neuron depends on the set of its connection weights. If the probability of firing should be high for some particular subset of the inputs, the weights representing words from those inputs must be high. The other weights should be relatively low due to the synaptic scaling mechanism. Based on this assumption, after training, for each output layer neuron we prune $\theta$ per cent of its incoming connections with the lowest weights. $\theta$ is a hyper parameter of the method empirically evaluated in the experimental section.

\section{Empirical evaluation and results comparison}\label{sec:experiment}
This section presents experimental evaluation of the method proposed. In subsection~\ref{sec:experimentplan} the technical issues related to the setup of experiment and implementation of the training and evaluation procedures are discussed. The final two subsections focus respectively on the experimental results and compare them with the literature.

\subsection{Experiment setup}\label{sec:experimentplan}
\subsubsection{Data set and implementation details}
The \textit{bydate} version\footnote{\url{http://qwone.com/~jason/20Newsgroups/20news-bydate.tar.gz}} of \textit{20 newsgroups} is a well known benchmark set in the text classification domain. The set contains newsgroups post related to different categories (topics) gathered from \textit{Usenet}, in which each category corresponds to one newsgroup. Categories are organised into a hierarchical structure with the main categories being \textit{computers}, \textit{recreation and entertainment}, \textit{science}, \textit{religion}, \textit{politics}, and \textit{forsale}. The corpus consists of $18\;846$ documents nearly equally distributed among twenty categories and explicitly divided into two subsets: the training one ($60\%$) and the test one ($40\%$).

The dynamics of the spiking neurons (including the plasticity mechanism) was implemented using the \textit{BRIAN 2} simulator~\cite{stimberg:good}. \textit{Scikit-learn Python library}\footnote{\url{https://scikit-learn.org/}}
was used for processing the text and creating the TF-IDF matrix.

%
\subsubsection{Training}
As mentioned in Section~\ref{learning_procedure} the training set was divided into $s=11$ subsets $u_{i}$ each of which, except for $u_{11}$, contained $1\;500$ documents. Division was performed randomly. Firstly the entire training set was shuffled, and then consecutively assigned to the subsets according to the resulting order with a $500$ document redundancy (overlap) between the neighbouring subsets, as described in Table~\ref{table_training_units}.

The overlap between subsequent subsets resulted from preliminary experiments which suggested that such an approach improves classification accuracy. While we found the concept of partial data overlap to be reasonably efficient, it by no means should be regarded as an optimal choice. The optimal division of data into training subsets remains an open question and a subject of our future research.
\begin{table}[h]
\centering
\caption{Division of the training set into subsets $u_{1}$-$u_{11}$.}
\begin{tabular}{lP{0.06\textwidth}P{0.06\textwidth}P{0.06\textwidth}P{0.06\textwidth}P{0.06\textwidth}P{0.06\textwidth}P{0.06\textwidth}P{0.06\textwidth}P{0.06\textwidth}P{0.06\textwidth}P{0.06\textwidth}P{0.06\textwidth}}
\hline
Subset  & $u_{1}$   & $u_{2}$   & $u_{3}$   & $u_{4}$   & $u_{5}$   & $u_{6}$   & $u_{7}$   & $u_{8}$   & $u_{9}$   & $u_{10}$   & $u_{11}$   \\ \hline
First index & 0    & 1000 & 2000 & 3000 & 4000 & 5000 & 6000 & 7000 & 8000 & 9000  & 10000 \\
Last index  & 1500 & 2500 & 3500 & 4500 & 5500 & 6500 & 7500 & 8500 & 9500 & 10500 & 11314 \\
Size & 1500 & 1500 & 1500 & 1500 & 1500 & 1500 & 1500 & 1500 & 1500 & 1500  & 1314  \\ \hline
\end{tabular}
\label{table_training_units}
\end{table}
%
%
%

\subsubsection{Evaluation procedure}
The outputs of all SNNs $H_i, i=1,\ldots,s$, i.e. spike-based encodings represented as sums of spikes per document were joined to form a single matrix (a final low-dimensional text representation) which was evaluated in the context of a classification task. The joined matrix of spike rates was used as an input to the Logistic Regression (LR)~\cite{Hosmer2000,James2013} classifier with \textit{accuracy} as the performance measure.

\subsection{Experimental results}\label{sec:expplanrealisation}
\begin{figure}[H]
\includegraphics[width=\textwidth]{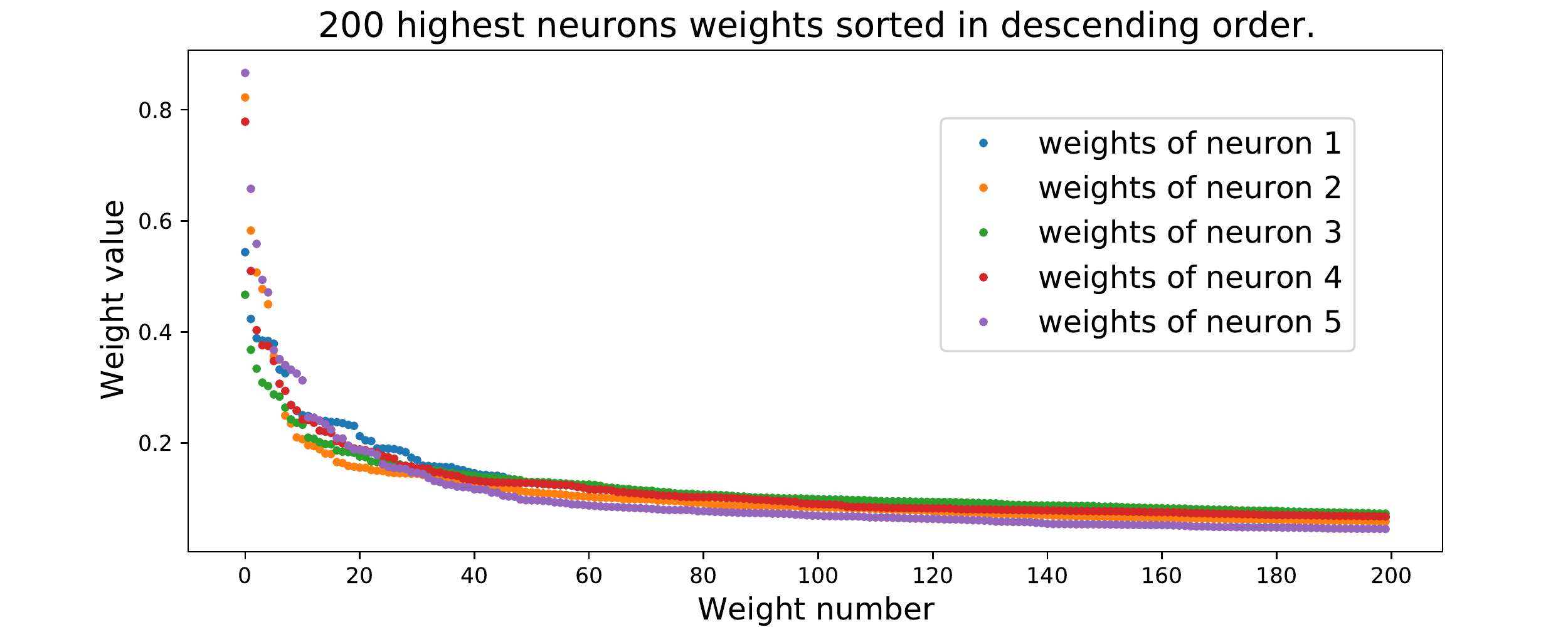}
\caption{Weights extracted from the encoder's neurons.}
\label{fig:extracted_weights}
\end{figure}

In the first experiment we looked more closely at the weights of neurons after training and the relationship between the inhibition mechanism and the quality/efficacy of resulting text representation. We trained eleven SNN encoders with $50$ neurons each according to the procedure presented above. After training, $5$ neurons from the first encoder ($H_1$) was randomly sampled and their weights were used for further analysis. Fig.~\ref{fig:extracted_weights} illustrates the highest $200$ weights sorted in descending order.

The weights of each neuron are presented with a different colour. The plots show that every neuron has a group of dominant connections represented by the weights with the highest values - the first several dozen connections. It means that each neuron will be activated more easily by the inputs that contain words corresponding to these weights. For example neuron $4$ will potentially produce more spikes for documents related to religion because its $10$ highest weights corresponds to words \textit{'jesus', 'god', 'paul', 'faith', 'law', 'christians', 'christ', 'sabbath', 'sin', 'jewish'}. A different behaviour is expected from neuron $2$ whose $10$ highest weights corresponds to words \textit{'drive', 'scsi', 'disk', 'hard', 'controller', 'ide', 'drives', 'help', 'mac', 'edu'}. This one will be more likely activated for computer related documents. On the other hand, not all neurons can be classified so easily. For instance $10$ highest weights of neuron $5$ are linked to words \textit{'cs', 'serial', 'ac', 'edu', 'key', 'bit', 'university', 'windows', 'caronni', 'uk'}, hence a designation of this neuron is less obvious. We have repeated the above sampling and weigh inspection procedure several times and the observations are qualitatively the same. For the sake of space savings we do not report them in detail.

\begin{figure}[!h]
\center
\includegraphics[width=0.8\textwidth]{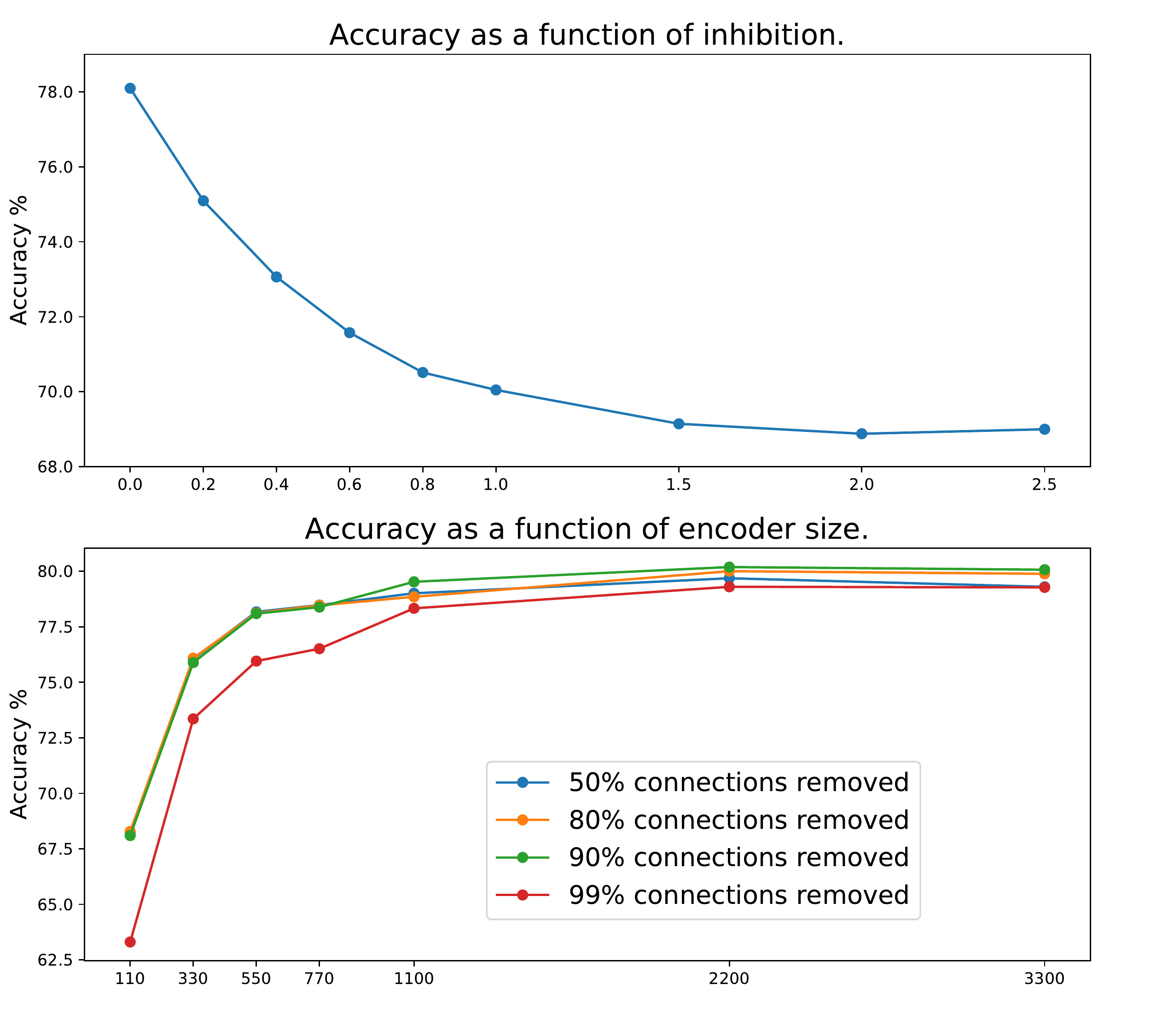}
\caption{Accuracy for various inhibition levels (top) and encoder sizes (bottom). }
\label{inhibiton}
\end{figure}

Hence, a question arises as to how well documents can be encoded with the use of neurons trained in the manner described above? Intuitively, in practice the quality of encoding may be related to the level of competition amongst neurons in the evaluation phase. If the inhibition value is kept high enough to satisfy WTA strategy then only a few neurons will be activated and the others will be immediately suppressed. This scenario will lead to highly sparse representations of the input documents, with just a few or (in extreme cases) only one neuron dominating the rest. Since differences between documents belonging to different classes may be subtle, such a sparse representation may not be the optimal setup. In order to check the influence of the inhibition level on the resulting spike-based representation we tested the performance of the trained SNNs $H_{1}$-$H_{11}$ for various inhibition levels by adjusting the value of the neurons' inhibitory synapses. The results are illustrated in Fig.~\ref{inhibiton}~(top).

Clearly the accuracy strongly depends on the inhibition level. The best outcomes ($\approx78\%$) are accomplished with inhibition set to $0$ and rapidly decrease along with the inhibition raise. For the inhibition values higher than $1.5$ the accuracy plot enters a plateau at the level of approximately $68\%$. The results show that the most effective representation of documents is generated with the absence of inhibition during the evaluation phase, i.e. when all neurons have the same chance of being activated and contribute to the document representation.

The second series of experiments aimed at exploring the relationship between the efficacy of document representation and the size of the encoders. Furthermore, the sensitivity of the trained encoders to \textit{connection pruning}, with respect to their efficiency, was verified. The results of both experiments are shown in the bottom plot of Fig.~\ref{inhibiton}. Seven encoders of various sizes (between $110$ and $3\;300$ neurons) were trained, and once the training was completed the before the \textit{connection pruning} procedure took place.

In the plot, four colored curves illustrate particular pruning scenarios and their impact on classification accuracy for various encoder sizes. $99\%$, $90\%$, $80\%$, and $50\%$ of the weakest weights were respectively removed in the four discussed cases. Overall, for smaller SNN encoders (between $110$ and $1\;100$ neurons) the accuracy rises rapidly along with the encoder size increase. For larger SSNs, changes in the accuracy are slower and for all four curves stay within the range $[77.5\%, 80.19\%]$.

In terms of the \textit{connection pruning} degree the biggest changes in accuracy (between $63\%$ and $79\%$) are observed when $99\%$ of connections have been deleted (the red curve). In particular, the results of the encoders with fewer than $1\;100$ neurons demonstrate that this range of pruning heavily affects classification accuracy. In larger networks additional neurons  compensate the features removed by the \textit{connection pruning} mechanism and the results are getting closer to other pruning setups.

Interestingly, for the networks smaller than $770$ neurons, the differences in accuracy between $50\%$, $80\%$, and $90\%$ pruning setups are negligible, which suggests that relatively high redundancy of connections still exist in the networks pruned in the range of $50\%$ to $80\%$. Apparently, retaining as few as $10\%$ of the weights does not impact the quality of representation and does not cause deterioration of results. This result well correlates with the outcomes of the weight analysis reported above and confirms that a meaningful subset of connections is sufficient for proper encoding the input. The best overall classification result ($80.19\%$) was achieved by the SNN encoder with $2\;200$ neurons and level of pruning set to $90\%$ (the green curve). It proves that SEM can effectively reduce dimensionality of the text input from initial $\approx 130\;000$ (the size of \textit{20 newsgroups} training vocabulary) to the size of $550-2\;200$, and maintain classification accuracy above 77.5\%.

\subsection{Results analysis. Comparison with the literature.}\label{sec:compmeth}
Since to our knowledge this paper presents the first attempt of using SNN architecture to text classification, in order to make some comparisons we selected results reported for other neural networks trained with similar input (\textit{document-term matrix}) and yielding low-dimensional text representation as the output. The results are presented in Table~\ref{tab:allResults}.
\begin{table}[!h]
  \centering
  \caption{Accuracy [\%] comparison for several low-dimensional text representation methods on \textit{bydate} version of \textit{20 newsgroups} data set.}
    \begin{tabular}{lc}
    \hline
          \textbf{Method} & \textbf{Accuracy} \\
    \hline
          SET (this paper) & \textbf{80.19} \\
          K-competitive Autoencoder for TExt (KATE)~\cite{Chen2017} & 76.14 \\
          Class Preserving Restricted Boltzmann Machine (CPr-RBM)~\cite{Hu2017} & 75.39 \\
          Variational Autoencoder~\cite{Chen2017} & 74.30 \\
    \hline
    \end{tabular}%
  \label{tab:allResults}%
\end{table}%
SET achieved $80.19\%$ accuracy and outperformed the remaining shallow approaches. While this result looks promising we believe that there is still room for improvement with further tuning of the method (in particular a division of samples into training subsets), as well as extension of the SNN encoder by adding more layers. Another interesting direction would be to learn semantic relevance between different words and documents~\cite{Gao2018,Zheng2016}.

\section{Conclusions}\label{sec:conclusions}

This work offers a novel approach to text representation relying on Spiking Neural Networks. Using the proposed low-dimensional text representation the LR classifier accomplished $80.19\%$ accuracy on a standard benchmark set (\textit{20 newsgroups bydate}) which is a leading result among shallow approaches relying on low-dimensional representations. 

We have also examined the influence of the inhibition mechanism and synaptic connections sparsity on the quality of the representation showing that (i) it is recommended that inhibition be disabled during the SNN evaluation phase, and (ii) pruning out as many as $90\%$ of connections with lowest weights did not affect the representation quality while heavily reducing the SNN computational complexity, i.e. the number of differential equations describing the network.

There are a few lines of potential improvement that we plan to explore in the further work. Most notably, we aim to expand the SNN encoder towards Deep SNN architecture by adding more layers of spiking neurons which should possibly allow to learn more detailed features of the input data.

\bibliographystyle{splncs04}
\bibliography{bibliography}

\end{document}